\documentclass{vgtc}                          

\ifpdf
  \pdfoutput=1\relax                   
  \usepackage{graphicx}                
  \DeclareGraphicsExtensions{.pdf,.png,.jpg,.jpeg} 
\else
  \usepackage{graphicx}                
  \DeclareGraphicsExtensions{.eps}     
\fi%

\graphicspath{{figures/}{pictures/}{images/}{./}} 

\usepackage{microtype}                 
\PassOptionsToPackage{warn}{textcomp}  
\usepackage{textcomp}                  
\usepackage{mathptmx}                  
\usepackage{times}                     
\usepackage{cite}                      
\usepackage{tabu}                      
\usepackage{booktabs}                  

\usepackage{xcolor}
\onlineid{0}

\vgtccategory{Research}

\vgtcinsertpkg

\title{The Development of Visualization Psychology Analysis Tools to Account for Trust }

\author{
Rita Borgo\thanks{e-mail: rita.borgo@kcl.ac.uk}\\ %
        \scriptsize Informatics Department, Kings College London, UK %
\and Darren J Edwards\thanks{e-mail: d.j.edwards@swansea.ac.uk\vspace{-0.2in}}\\ %
     \scriptsize Department of Public Health, Policy, and Social Sciences, Swansea University, UK}

\abstract{Defining trust is an important endeavor given its applicability to assessing public mood to much of the innovation in the newly formed autonomous industry, such as artificial intelligence (AI), medical bots, drones, autonomous vehicles, and smart factories \cite{Yin:2018}.  Through developing a reliable index or means to measure trust, this may have wide impact from fostering acceptance and adoption of smart systems to informing policy makers about the public atmosphere and willingness to adopt innovate change, and has been identified as an important indicator in a recent UK policy brief \cite{Edwards:2019}. In this paper, we reflect on the importance and potential impact of developing \emph{Visualization Psychology} in the context of solving definitions and policy decision making problems for complex constructs such as ``trust''.%
} 

\begin{document}


\firstsection{Introduction}

\maketitle

Trust has been studied from various perspectives such as psychology, sociology, philosophy, political science, economics, and human factors \cite{Hoff:2015}.  However, defining trust is not straightforward as there may be many situational and contextual factors which need to be considered.  For example, how does interpersonal trust relate or differ to trust in artificial settings? A very topical example being human-machine partnership involving intelligent and autonomous systems. 
The literature~\cite{Mayer:1995} identifies three general levels to define the bases of trust: ability, integrity, and benevolence. Ability is defined as the set of skills, competencies, and characteristics that enable a trustee to influence the trust domain.
Integrity is defined as the degree to which trustee and trustor
adheres to a set of principles acceptable to both parties. Benevolence is defined as the extent to which the intents and motivations of trustee and trustor are aligned.
Further studies \cite{Lee:2004} have suggested that fully mature (intimate, for example) relationships are based also on faith (e.g., fidelity) as a natural development of benevolence, a trait which would assume different connotations in the context of trust in autonomous systems.  Hoff and Bashir \cite{Hoff:2015} suggest in fact that trust within the context of interpersonal relations bears similarities to that of trust in automation as they both share a willingness to rely on the trustee under uncertainty.
Hoff and Bashir \cite{Hoff:2015} also suggest three layers that may define human-automation trust, these being: (1) dispositional trust; (2) situational trust; and (3) learned trust.  
Trust however is a dynamic and personal phenomena and as such its definition include many more layers~\cite{Barber:1983} whose investigation goes beyond the scope and space of this paper.


We wish instead to explore how Visualization has been employed as one of the means to enable the understanding of trust when problem domain and tentative solutions carry a high level of complexity and associated risk.
We have seen the emergence of fields such as Visualization for Explainable AI where visual layouts are  used to support the decision process by unpacking the inner workings of black-box techniques, supporting dynamic exploration of complex decision trees, providing different level of abstractions tailored to both task and end-user abilities. Yet visualization is not a standalone tool as the challenge imposed by trying to define, measure, influence, nudge ``trust'' is a multi-disciplinary one~\cite{Vandenhoven:2015}.

\begin{figure*}[t]
    \centering
    \includegraphics[width =\textwidth]{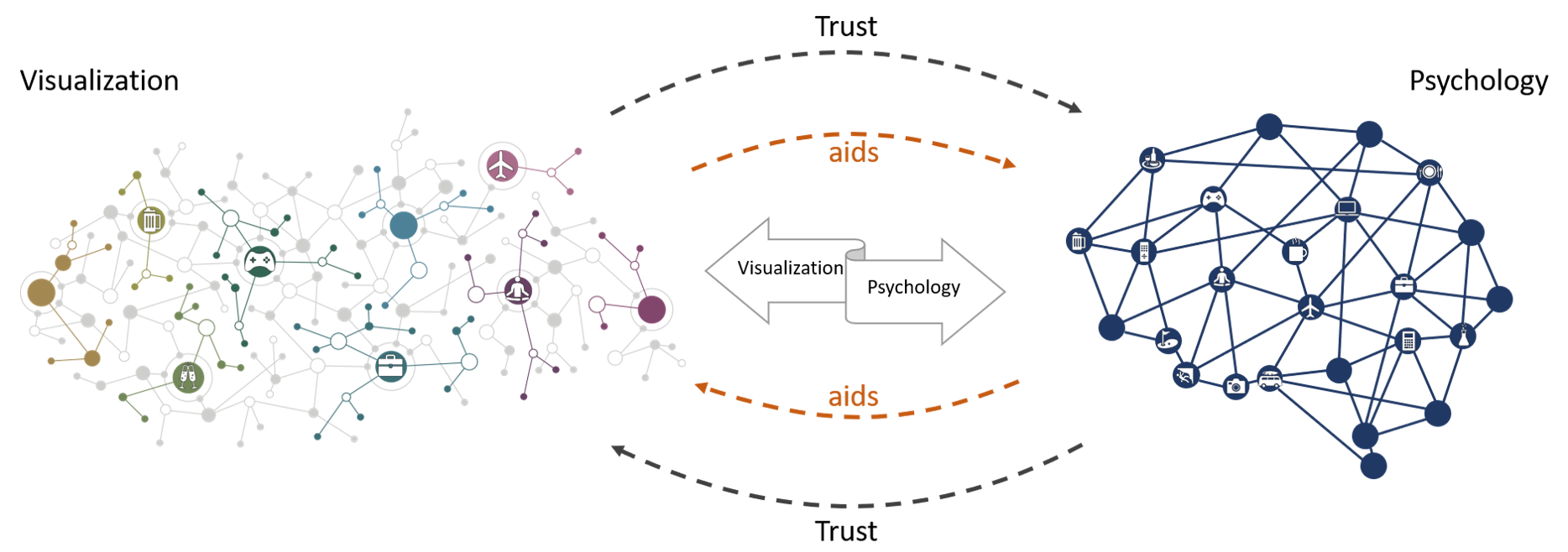}
    \caption{Visualization Psychology for Trust.}
    \label{fig:VisPych}
\end{figure*}

In this paper we look at \emph{Visualization Psychology} as a candidate subject to address the challenge of defining tools and metrics to asses trust through the means of visualization and vice-versa. We see this as a two-fold challenge, as depicted in Fig.~\ref{fig:VisPych}: (1) Can visualization help to explain the psychological and sociological phenomena of trust? (2) Can psychology help to identify what makes a visualization trustworthy from the user's perspective and/or influence their propensity to trust especially in the domain of human-machine interaction?  

\section{Visualization Psychology - Visualization as Tools}
Visualization psychology has been developed as a new branch of applied psychology, while its scope is yet to be defined one area of application is the study of the effects of visualization on different data intelligence workflows. When faced with the problem of defining trust in the context of social, cultural, and environmental factors (such as social media influence), there exist areas of visualization which provide effective means of analysis that benefit the exploration and understanding of complex psychological phenomena.
\paragraph{Network Visualization.} 
Given the hierarchical and graph based nature of intelligence workflows \textit{Network visualization} is a very promising area~\cite{Radicchi:2004}. Community network analysis for example, which is typically based on graph theory, is useful to understand the contextual network and important features of trust, as it can plot the relationships between individuals, community groups such as peers, media, online influences, cultural influences etc. This allows a mapping of the important constructs that make up a ``trust network'' for a particular individual. 
Network analysis uses several mining metrics, with one example being Cohen’s $ \kappa $, a similarity measure for categorical data. Cohen’s $ \kappa $ provides a concrete summative result enriched by resultant networks displayed using node-link diagrams \cite{Hoffman:2017} for explaining the analytical results. Metrics like Cohen’s $ \kappa $ and relative graph based explanations are being employed as support to more qualitative and subjective results gathered through empirical studies. This for instance is useful in explaining a complex and contextually based construct such as ``trust'' in a particular setting such as AI in automated industry. Similarly it is as effective for the unpacking of a complex definition within its unique set of situational circumstances. 

Psychology research has shown that trust has mainly two bases: trustworthiness - the extent to which a trustee is competent, honest, and has goodwill toward the trustor - and trust propensity - a stable trait reflecting the trustor's generalized belief that others can be trusted~\cite{Baer:2018}. This raises two interesting questions: which aspects of the social context might favour the establishment of trustworthiness? What aspects of the social context might cause people to experience fluctuations in their trust propensity?
In the context of autonomous systems Visualization Psychology can propose models which may be usefully applied to formalise an answer. An example is the Elaboration Likelihood model (ELM) which suggests that users elaborate information deeply to help them decide whether the object, person, or activity should be judged as trustworthy \cite{Kong:2019}. ELM relies on the definition of two routes to persuasion, one central one peripheral. Different stressing factors define the choice between either one of the routes. 
Factors that may be as important are user experience, which can increase trust if positive \cite{Costante:2011}, the credibility of information \cite{Kong:2019}, a self-calibrated degree of confidence. An ELM diagram can therefore easily be converted to a dynamic weighted graphs with multiple layers of encoded information.

\section{Visualization Psychology -  Tools for Visualization}

Visualization psychology can help define visualization supported metrics to identify trust enablers, their interconnection and influence. When the problem domain increases, sophisticated visualization tools may be usefully applied to include multi-layered information such as context (both social and situational), community and other sources of influence within the same visual layout.  
Parametric visual encoding \cite{Borgo:2013} for example fall into this category in that they allow for both visualization summaries and contextual as well as semantic zooming. Information visualization literature has leveraged tools from different areas of psychology (cognitive, behavioural etc.) to characterise strengths and limitations of complex visual layouts and to help define guidelines.
In the context of trust increase in visual and ontological complexity of a visualization conversely introduces another question that visualization psychology may address, that is: the definition of metrics and measures for identifying enablers of trust in information visualization techniques themselves.
Mayer et al. survey on trust in Information Visualization~\cite{Mayr:2019} highlights a series of important challenges which emphasize how the answer to the question ``what makes a visualization trustworthy'' is beyond cognition and perception alone. The analytical process to reach trust may rely on both an articulate thought process as well as the use of superficial but straightforward cues that act as indicators and triggers for trust~\cite{Dasgupta:2016}.
Mayr et al.~\cite{Mayr:2019} focus the attention on two aspects of trust in visualization: trust in the data and trust in the visual representation of the data.
Clarity, amount of information depicted and disclosed in a visualization are factors that may influence trust \cite{Xiong:2019} while effects of both visual complexity and information complexity remain still undetermined. There are however (yet) no unique indicators to determine enablers of trust in the quality of information~\cite{Lucassen:2011} or in the actual visualization itself~\cite{Xiong:2019, Doerk:2013}. 
These challenges seem to support Visualization Psychology as an appropriate subject discipline to contribute to the identification of which and how design cues influence trust in information visualization.


\section{Philosophical Considerations}
Visualization Psychology draws schools of thoughts from two distinct subject areas, that is psychology and visualization. Psychology is largely concerned with explaining phenomenon that exists relating to mind or behavior. Visualization, on the other hand, is concerned with depicting information in ways which are easily accessible to the cognitive system~\cite{Edwards:2020}. Some concepts such as trust are not easily definable, so the application of visualization psychology in this instance is ideal. However, greater ontological accounting of these approaches in the various domains, may be required to substantiate the methodological need for this approach.

\section{Conclusions}
Trust within Visualization Psychology as a new discipline, may depend on its relative success to explain psychological phenomenon such as the ``defining trust problem'' and potentially many others. 
It is highly desirable to visualize trust within the context of specific situational factors. Visualization Psychology approaches reinforced by visualization, such as community network analysis for example, may be one mean of discovering the important components of trust within context.  
Orthogonal to this is the question on whether experts as well as lay users can completely trust the visual representation of information.
This has become particularly relevant when the analytical process and decision making are instances of mixed-initiative systems where human-machine partnership is key.
Visualization can potentially support transparency at the analysis process level as well as the final output~\cite{Dasgupta:2016}, yet transparency may not be sufficient to effect trust. Visualization Psychology appears as a promising subject discipline to support and foster the development of the fundamentals of trust across both Psychology and Visualization. Psychology as a discipline is already embedded in the visualization research community through evaluation and empirical studies, therefore is there really a need for a new research area like Visualization Psychology or what already exists within the Visualization community is sufficient? While Visualization Psychology could as well act a ``container''  for the formalization of expertise and knowledge acquired through the synergy of these two disciplines are there further synergies not yet explored? These and other questions are the subject of our viewpoint contribution to the workshop.



\bibliographystyle{abbrv-doi}

\bibliography{template}
\end{document}